\def\year{2022}\relax
\documentclass[letterpaper]{article} 
\usepackage{aaai22}  
\usepackage{times}  
\usepackage{helvet}  
\usepackage{courier}  
\usepackage[hyphens]{url}  
\usepackage{graphicx} 
\usepackage{natbib}  
\usepackage{caption} 
\DeclareCaptionStyle{ruled}{labelfont=normalfont,labelsep=colon,strut=off} 
\usepackage{amsmath,amsthm}
\usepackage{amssymb}
\usepackage{bbm}
\usepackage{amsfonts}
\usepackage{graphicx}
\usepackage{subfigure}
\graphicspath{{graphics/}}
\usepackage[ruled,vlined]{algorithm2e}
\usepackage{algorithmic}
\usepackage{tabularx,booktabs}
\usepackage{tikz}

\usepackage{xcolor}
\definecolor{darkblue}{rgb}{0.0,0.5,0.5}
\usepackage{booktabs,caption}
\usepackage{multirow}
\usepackage{lscape}
\usepackage{epstopdf}
\usepackage{lineno}
\usepackage{subfigure}

  \usepackage{fixmath}

\renewcommand{\vec}[1]{\mathbold{#1}}
 
\begin{document}
%
\title{Individual Mobility Prediction via Attentive Marked Temporal Point Processes}
\author{
    Yuankai Wu, Zhanhong Cheng, Lijun Sun\thanks{Corresponding author.}
}

\affiliations{
    McGill University, Montreal, QC, Canada\\
    yuankai.wu@mail.mcgill.ca, zhanhong.cheng@mail.mcgill.ca, lijun.sun@mcgill.ca
}
\maketitle
\begin{abstract}
Individual mobility prediction is an essential task for transportation demand management and traffic system operation. There exist a large body of works on modeling location sequence and predicting the next location of users; however, little attention is paid to the prediction of the next trip, which is governed by the strong spatiotemporal dependencies between diverse attributes, including trip start time $t$, origin $o$, and destination $d$. To fill this gap, in this paper we propose a novel point process-based model---Attentive Marked temporal point processes (AMTPP)---to model human mobility and predict the whole trip $(t,o,d)$ in a joint manner. To encode the influence of history trips, AMTPP employs the self-attention mechanism with a carefully designed positional embedding to capture the daily/weekly periodicity and regularity in individual travel behavior. Given the unique peaked nature of inter-event time in human behavior, we use an asymmetric log-Laplace mixture distribution to precisely model the distribution of trip start time $t$. Furthermore, an origin-destination (OD) matrix learning block is developed to model the relationship between every origin and destination pair. Experimental results on two large metro trip datasets demonstrate the superior performance of AMTPP.
\end{abstract}

\section{Introduction}

Understanding and predicting human mobility is critical to numerous smart city applications, such as intelligent transportation systems, location-based services, personalized and contextual advertisement,  built environment and infrastructure planning, to name but a few \citep{wang2019urban}. Thanks to recent advances in information and communications technology (ICT), large quantities of digital traces recording individual mobility (e.g., mobile phone data and transit smart card data) are generated continuously with a high spatiotemporal resolution, bringing significant advances in individual mobility/trajectory modeling and forecasting. Essentially, individual mobility data can be organized into two types: time-stamped location sequence $\{(t_i,l_i)\}_{i=1}^n$ ($l_i$ is user location at time $t_i$) and trip/activity sequence $\{(t_i,o_i,d_i)\}_{i=1}^n$ (leaving origin location $o_i$ for destination $d_i$ at time $t_i$). The majority of research focuses on predicting the next location of a user, while there is relatively little work on predicting the next trip. Compared with time-stamped location records, the multi-dimensional trip records encode much richer information about individual mobility, in particular characterizing the complex interactions/dependencies between location and time. As a result, trip records are more useful than trajectories for many downstream applications such as demand management and urban planning \citep{zhao2018individual,wang2019origin}. However, predicting individual trip $(t, o, d)$ is very challenging due to its multi-dimensional nature. A common practice in transportation system analysis is to model locations using $S$ categorical variables (e.g., traffic zones or train stations). In this case, the size of the spatial domain of $o\times d$ will be $S\times S$. The dependency structure will become even more complex if we further consider trip time $t$ and incorporate other attributes such as travel mode (e.g., by car, bike, or bus) and activity/travel purpose (e.g., for work/school/leisure).

\begin{figure}[!b]
\centering
\includegraphics[width=0.95\columnwidth]{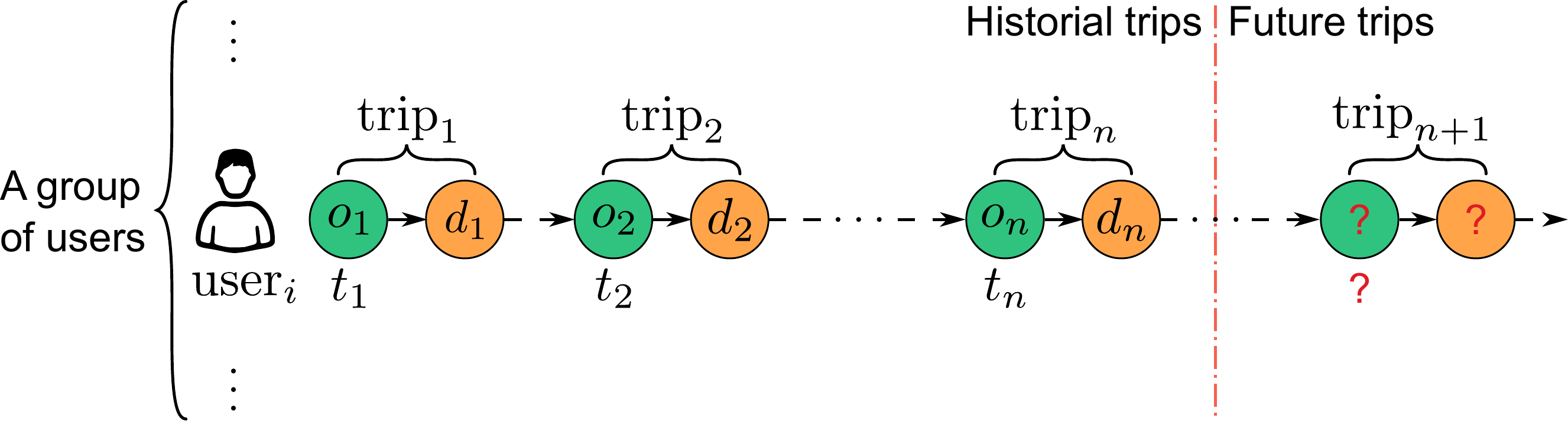}
\caption{Formulation of the individual trip prediction problem. We consider historical user mobility data as a list of trips, with trip $i$ starting at location $o_i$ at time $t_i$ and ending at location $d_i$. Our goal is to predict the next/future trip based on a user's historical trip records.}
\label{Fig:1}
\end{figure}

The main objective of this paper is to develop a novel approach to simultaneously predict the time $t_{n+1}$, origin $o_{n+1}$ and destination $d_{n+1}$ of the next trip given a user's trip history $\{t_i, o_i, d_i\}^n_{i=1}$ (see Figure~\ref{Fig:1}). Note that $d_{i}$ can be different from $o_{i+1}$ as the available trajectory data from a particular source is often incomplete. If we consider time $t$ a continuous variable and both origin $o$ and destination $d$ categorical variables, the domain of the $(n+1)$th trip variable will be $\left[t_{n},+\infty\right)\times \{1,\ldots,S\} \times \{1,\ldots,S\}$. Existing studies mainly model such event sequences in two ways: (1) hidden Markov models (HMM) and (2) marked temporal point processes (TPPs). A main limitation of HMM is that it employs a regular and discrete representation of time and thus fails to model events that occur in continuous time \citep{yin2017generative}. As a more general mathematical framework for event sequence modeling, a TPP is typically represented as $\mathcal{T}_t = \{(t_1, m_1), \ldots, (t_n, m_n) \}$ of strictly increasing time with a marker $m$ (also referred to as event type) from a small discrete label space.

Recent research has shown increasing interest in coupling TPPs with deep neural networks to jointly modeling event markers and timings \cite[see e.g.,][]{du2016recurrent,mei2017neural}. However, these approaches cannot be used to model human mobility data or trip sequences due to the following challenges. First, the marker of a trip event is multi-dimensional with at least two attributes---origin and destination, and for trip data there exist complex spatiotemporal and higher-order dependencies among timing, origin and destination. Not only do we have to deal with a large marker space of $S\times S$, but $S$ itself is also large at the scale of $\sim$100. To model timing-marker relationship, several studies assume static correlation among the timings of each marker \citep{mei2017neural,zhang2020self}.  Nevertheless, the learning parameters of these works become extremely high when the marker dimension increases. Second, most existing methods either adopt a temporal intensity function that decays exponentially following the assumption of Hawkes process or design a mixture distribution with learnable parameters; however, they are insufficient and unsuitable to model and characterize travel behavior, which is not only self-exciting but also of strong rhythms, periodicity and regularities \citep{sun2013understanding}. For example, a regular commuter makes two trips every day---going to work and returning home, and the trip  timings are mainly determined by his/her daily working schedule \citep{cheng2021incorporating}. Moreover, the inter-trip time here has a physical meaning---duration for the work activity, which follows a distribution with a strong peak at 8-10 hours.

In this work, we propose a novel approach named Attentive Marked Temporal Point Processes (AMTPP) to address the aforementioned challenges. First, we employ the self-attention mechanism to measure
the influence of past trips on the next trip. To fully utilize the temporal information and behavior periodicity, we develop a novel positional embedding method to effectively characterize temporal information. Second, we propose to use the mixture of Asymmetric Log-Laplace (ALL) mixture distribution to model inter-event times. The ALL distribution is flexible to learn the skewness and peak of activity duration and by using the mixture formulation we can also characterize the rhythms and regularities of travel behavior. Third, we introduce a novel origin-destination (OD) matrix learning block, in which a dynamic OD relationship matrix can be learned from the data in an end-to-end manner. We conduct extensive experiments on two real-world data sets to evaluate the proposed AMTPP framework, and our results confirm the superiority of AMTPP in reconstructing complex trip distributions as well as performing accurate stochastic forecasting. This paper brings two problem-specific contributions in using TPP to model human mobility and activity data: (1) we use time embedding to capture the complex periodical patterns rooted in human behavior and introduce asymmetric log-Laplace mixture distribution to model the conditional intensity with prominent peaks; (2) we design an OD matrix learning block to model $o$ and $d$ as two coupled markers, which can effectively uncover the underlying OD distribution from the trip data without any prior knowledge.

\section{Related Work}

\subsection{Individual Trip Modeling}

Individual mobility is governed by strong spatial and temporal regularities \citep{gonzalez2008understanding}. Predicting a person's next location from a sequence of his/her past location records has been extensively studied in the literature. For example, \citet{gambs2012next} show that a simple Markov model can accurately predict the next location. Mixed Markov models \citep{asahara2011pedestrian} are also introduced to capture the heterogeneity among individuals. Due to the great representation power of deep learning, RNN-based models have been increasingly used for next location prediction in recent studies \citep{feng2018deepmove}. Besides answering the ``where'' problem, many studies extend the scope to also predict ``when'' the next movement will happen, including continuous-time Markov model \citep{gidofalvi2012and}, temporal point processes \citep{du2016recurrent}, and RNN-based models \citep{sun2021joint}. An accurate mobility prediction requires continuous observation of individuals' complete trajectories, which is often sensitive and unrealistic to obtain. Instead of modeling continuous location sequences, a few studies have focused on the trip-based mobility representation and developed models to predict the entire trip/activity---including time, origin, destination, and activity duration---in a partially observed system based on $T$-gram approach \citep{zhao2018individual} and input-output HMM  \citep{mo2021individual}.

\subsection{Marked Temporal Point Processes (TPPs)}

A TPP is fully specified by a conditional intensity function, which describes the probability of an event occurring at any time given the history. Traditional TPPs often specify a simple parametric intensity function \citep{hawkes1971spectra}, which may lead to poor results because of its limited capacity in modeling data with increased complexity. Several recent studies develop develop neural TPPs to address this issue, by coupling TPPs with deep neural networks to achieve strong representation power. For example, \citet{du2016recurrent,mei2017neural,omi2019fully, mehrasa2019variational} use RNNs to encode the history of TPPs and then approximate the intensity function by some parametric forms. However, \citet{shchur2019intensity} argues that intensity function-based deep learning approaches can not achieve flexibility, efficiency, and ease-to-use simultaneously. Alternatively, the authors propose an intensity-free learning model of TPPs using log-Normal mixture distribution to represent complex probability distributions in a flexible way. A unique advantage of this approach is that it has closed-form expressions of likelihood. \citet{zhang2020self} proposes the Self-Attentive Hawkes Process (SAHP) model, which uses the self-attention mechanism to summarize historical events. The attention mechanism-based approach has demonstrated superior performance in capturing long-range dependency compared to RNN models, and hence SAHP outperforms other RNN baselines on cases with low-dimensional marks. However, the learning of SAHP is achieved by Monte Carlo integration since the model uses a constant intensity function without a closed-form likelihood. For TPPs with a single high-dimensional marker (e.g., $S$ is large), several solutions have been proposed based on graph structure learning  \citep{xiao2018learning} and generative adversarial networks \citep{wu2019learning}. The multi-dimensional marker problem has been considered in a few studies. For traditional TPPs with coupled markers, \citet{wu2018decoupled} proposes to factorize multiple coupled marks into a small set of interdependent processes to reduce model complexity at the cost of losing model expressiveness. Despite the significant progress in coupling deep learning and TPPs, none of the existing works can be used for the trip prediction problem. The temporal processes of human mobility are strongly governed by the regularity and routine in human behavior, of which the intensity function is time-sensitive and substantially different from traditional TPPs. Moveover, the high- and multi-dimensional ($S\times S$ with large $S$) markers also pose a new challenge to neural TPPs.

\section{Problem Description}

We focus on predicting the next trip of an individual based on his/her trip history. We denote the trip sequence of a user $u$ up to time $t$ by
\begin{equation}
    \mathcal{H}^u_t = \{(t^u_1, o^u_1, d^u_1), \ldots, (t^u_{n_u}, o^u_{n_u}, d^u_{n_u}) :t^u_{n_u} \leq t\},
\end{equation}
where $n_u$ is the number of trips in the sequence; $t^{u}_{i}$, $o^u_i$, and $d^u_i$ represent the departure time, origin location, and destination location of the $i$-th trip, respectively. Departure time is a continuous variable, and we consider origin and destination two discrete attributes for the trip. Note that the dimension of features can be further increased by including other attributes such as travel mode and travel purpose. We omit the index $u$ in the following of this paper, since our model is universal to all users. The timing $t_i$ in a trip sequence $\mathcal{H}_t$ can also be represented by a sequence of strictly positive inter-event time intervals $\tau_i = t_i - t_{i-1} \in \mathbb{R^{+}}$. For mobility modeling, the term $\tau_i$ can be roughly considered the activity duration between the $(i-1)$th trip and the $i$th trip. The two representations using $t_i$ and $\tau_i$ are isomorphic, and we will use them interchangeably throughout the paper.

The goal of this study is to estimate the probability distribution of the next trip $(\tau_{n+1}, o_{n+1}, d_{n+1})$ given the history sequence $\mathcal{H}_{t_n}$:
\begin{equation}
    p^{*}(\tau_{n+1}, o_{n+1}, d_{n+1}) = p \left(\tau_{n+1}, o_{n+1}, d_{n+1} \mid \mathcal{H}_{t_n}\right),
\end{equation}
where we use symbol $*$ to denote that $p^{*}$ is conditioned on $\mathcal{H}_{t_n}$, and the prediction is updated every time when we observe a new trip. Note that in this paper we model trip sequence pertaining to a particular mode (e.g., train). If the complete mobility information of an individual is accessible, we expect the next origin to be the same as the last destination ($o_{n+1} = d_n$), and in this case the problem reduces to a TPP with a single high-dimensional marker coded using either origin or destination.

\section{Methodology}

\begin{figure}[!t]
\centering
\subfigure[Self-attention based history encoder for AMTPP.]{%
\label{fig:2first}%
\includegraphics[width=0.9\columnwidth]{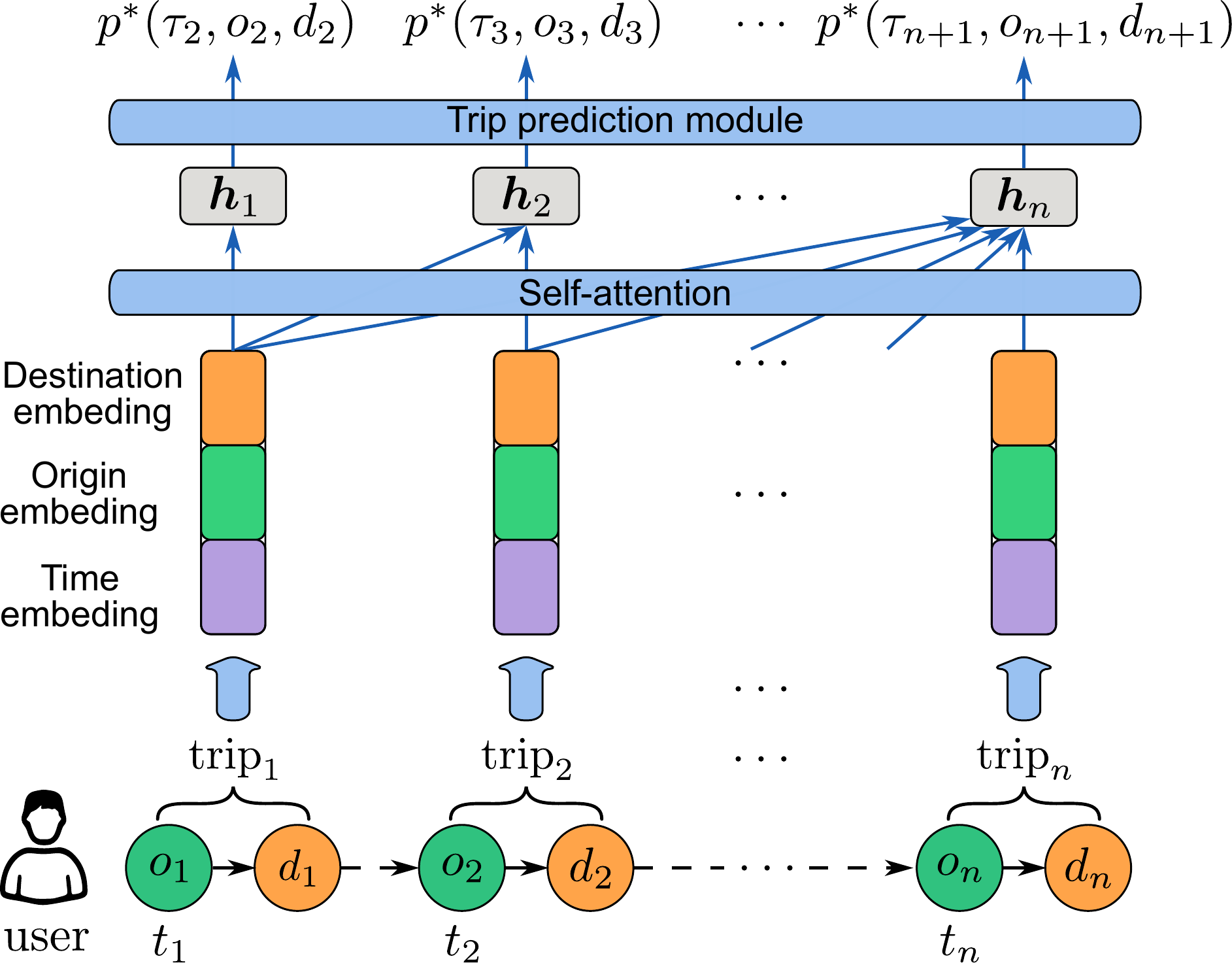}}
\subfigure[Trip prediction module of AMTPP, which generates $p^{*}(\tau_{n+1})$, $p^{*}(o_{n+1})$ and $p^{*}(d_{n+1})$ using the output of history encoder.]{%
\label{fig:2second}%
\includegraphics[width=0.9\columnwidth]{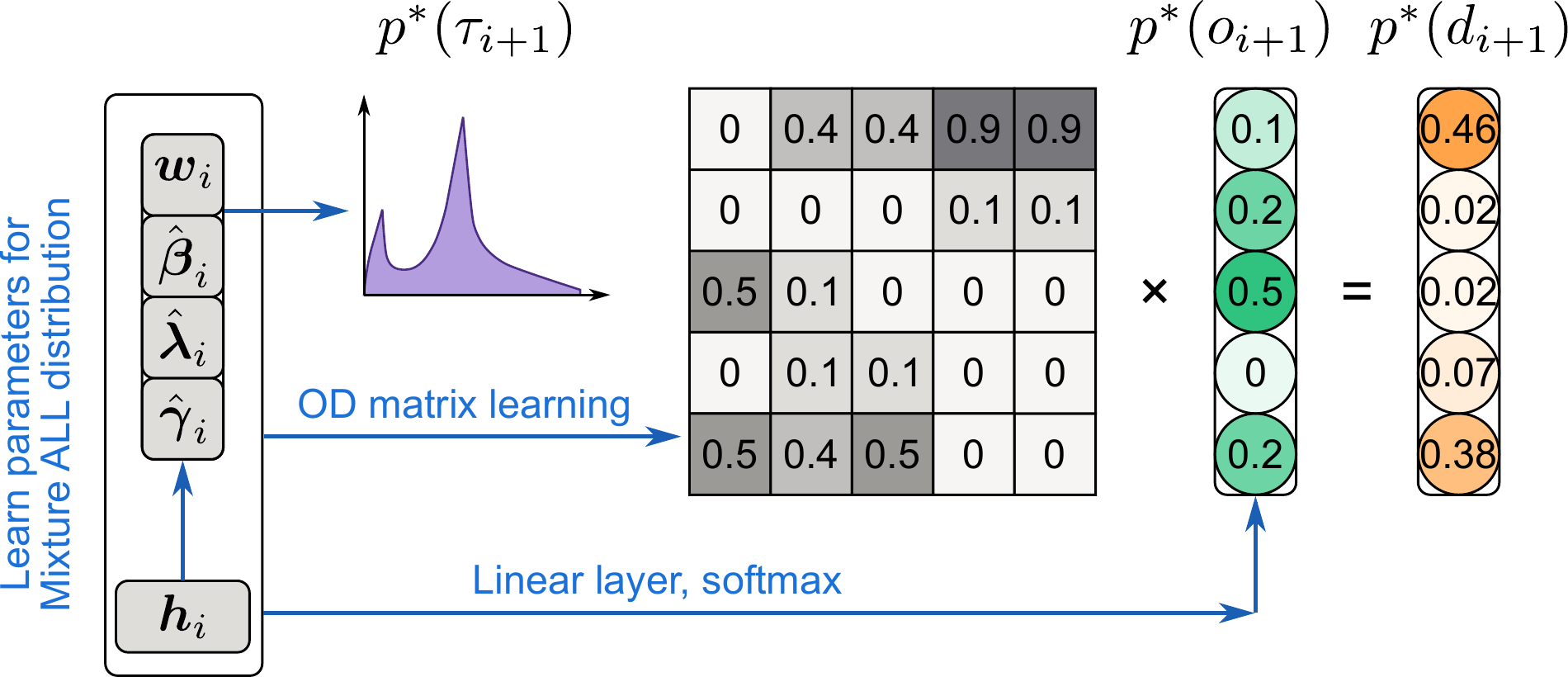}}%
\caption{Model Architecture of AMTPP. }
\label{fig:2}
\end{figure}

Figure~\ref{fig:2} illustrates the architecture of AMTPP designed for individual mobility prediction. In particular, we use a self-attention module to obtain the history encoding $\left\{\vec{h}_0, \vec{h}_1, \ldots, \vec{h}_n \right\}$ from a user's trip sequence, and model intensity distribution $p^{*}(\tau_{n+1})$ using an asymmetrical log-Laplace mixture, of which parameters $\theta_{n+1}$ are conditioned on $\vec{h}_n$. The distribution of next origin $p^{*}(o_{n+1})$ is conditioned on both history encoding $\vec{h}_n$ and parameters of $p^{*}(\tau_{n+1})$. An OD matrix learning block is developed to learn the OD transition. Finally, the distribution $p^{*}(d_{n+1})$ is obtained by multiplying $p^{*}(o_{n+1})$ with the OD transition matrix.

\subsection{Self-attention Encoder}

The first step of AMTPP is to obtain a representative embedding from $t_n$, $o_n$ and $d_n$ of the $n$-th trip. In AMTPP, the embedding $\vec{e}_n$ contains three parts:
\begin{equation}
    \vec{{e}}_n = \operatorname{concat}(\vec{{emb}}^{t}_n, \vec{{emb}}^{o}_n, \vec{{emb}}^{d}_n),
\end{equation}
where $\vec{{emb}}^{t}_n$, $\vec{{emb}}^{o}_n$ and $\vec{{emb}}^{d}_n$ represents embeddings for time, origin, and destination, respectively. As we apply the same operation for all users, the user index is omitted.

To account for the periodicity and rhythms of travel behavior, we introduce positional encoding \citep{vaswani2017attention} for the hour of day:
\begin{equation}
\begin{split}
\vec{pe}^h_n(pos_n, 2i) &= \sin\left(pos_n/L_{pos}^{{2i}/{J}}\right), \\
\vec{pe}^h_n(pos_n, 2i+1) &= \cos \left(pos_n/L_{pos}^{{2i}/{J}}\right),
\end{split}
\end{equation}
where $pos_n\in \{0, 1,\ldots,23\}$ is the hour of day of the $n$-th trip, $J$ is the total dimension of the embedding, $i \in \{1, \ldots, \lfloor J/2 \rfloor \}$, and $L_{pos}$ is the scaling factor. Although a fixed $L_{pos}$ (e.g, 10,000) is often selected in defining positional embedding, we treat it as a parameter because it could be beneficial for general approximation \citep{wang2020position}.
Similarly, we also introduce positional encoding $\vec{pe}^w_n$ for the day of week. Combining with the time lag $\tau_n$, the final time embedding $\vec{{emb}}^{t}_n$ is given by
\begin{equation} \label{eq:time_embedding}
\vec{{emb}}^{t}_n = \operatorname{concat}( \vec{pe}^w_n, \vec{pe}^h_n, \tau_n).
\end{equation}

As origin $o_n$ and destination $d_n$ contain both contextual and spatial information, we use a learnable embedding layer to encode origin and destination information:
\begin{equation}
\begin{split}
    \vec{{emb}}^o_n = \operatorname{concat}( W^o_{em} \vec{\hat{o}}_n + \vec{b}^o_{em}, \vec{p}_n^o), \\
    \vec{{emb}}^d_n = \operatorname{concat}( W^d_{em} \vec{\hat{d}}_n + \vec{b}^d_{em}, \vec{p}_n^d),
\end{split}
\end{equation}
where $W^o_{em} \in \mathbb {R}^{J_o \times S}$, $W^d_{em} \in \mathbb {R}^{J_d \times S}$, $\vec{b}^o_{em} \in \mathbb {R}^{J_o}$ and $\vec{b}^d_{em} \in \mathbb {R}^{J_d}$ are parameters, $S$ is the total number of locations, $J_o$ and $J_d$ are the dimension of the origin and destination embedding vector, and $\vec{p}^o_n$ and $\vec{p}^d_n$ are vectors containing additional information (e.g., point of interest) of $o_n$ and $d_n$.

Given a series of historical events and their associated embedding $E_n = [\vec{{e}}_1, \vec{{e}}_2, \ldots \vec{{e}}_n]^{\top} \in \mathbb{R}^{n \times J}$ until $t_n$, we compute the hidden state of the $n$-th trip by employing the multi-head self-attention mechanism. The multi-head self-attention layer transforms the embedding $E$ into $L$ distinct query matrices $Q_l = E W^Q_l$, key matrices $K_l = E W^K_l$, and value matrices $V_l = E W^v_l$, respectively, $l = 1,2, \ldots,L$. Here, $W^Q_l, W^K_l \in \mathbb{R}^{J \times c_k}$ and $W^V_l \in \mathbb{R}^{J \times c_v}$ are learnable parameters. Note that we omit the trip index in $E_n$, as the same operation is applied for different $n$. After these linear projections, the scaled dot-product attention computes a sequence of vector outputs:
\begin{equation}
     {{\operatorname{Att}}(Q_l,K_l,V_l)={\operatorname{softmax}}\left({\frac {Q_lK_l^{\mathrm {T}  }}{\sqrt {d_{k}}}\cdot M}\right)V_l},
     \label{eq:att}
\end{equation}
where a mask matrix $M \in \mathbb{R}^{n \times n}$ is applied to filter out rightward attention by setting all upper triangular
elements to $-\infty$ to avoid future information leakage. We then apply a multi-head attention layer:
\begin{align}
\begin{split}
    H &= \operatorname{gelu}(\operatorname{concat}(\text{head}_1,\ldots, \text{head}_L) W^O), \\
    \text{head}_l &= \operatorname{Att}(E W^Q_l, E W^K_l, E W^V_l),
\end{split}
\end{align}
where $W_O \in \mathbb{R}^{L \cdot c_v \times c_{\text{model}}}$, $c_{\text{model}}$ is the size of output feature, function $\operatorname{gelu}$ represents the Gaussian Error Linear Unit for nonlinear activations, and $H_n = [\vec{{h}}_1, \vec{{h}}_2, \ldots \vec{{h}}_n]^{\top} \in \mathbb{R}^{n \times c_\text{model}}$ is the output. Since we use a mask matrix $M$ to filter out all values in the input sequence that correspond to future events, the encoding $\vec{{h}}_i$ in $H$ is only based on itself and its history.

\subsection{Asymmetrical Log-Laplace Mixture for Inter-trip Time}

In this section, we show how to model the conditional probability distribution of inter-trip time $p_{\theta}\left(\tau_{n+1}  \mid \vec{h}_n\right)$ with deep neural networks parameterized by $\theta$. \citet{shchur2019intensity} suggests that it is more convenient to directly modeling the conditional distribution of inter-event time than modeling the intensity function. The authors use a log-normal mixture model to estimate the conditional probability function $p_{\theta}(\tau_{n+1} \mid \vec{h}_n)$ for TPPs. However, this log-normal formulation is inappropriate to model inter-trip intervals, which have a physical meaning of activity duration (e.g., time spent at home/work/school/leisure), and the conditional distributions often show prominent peaks \citep{sun2013understanding,cheng2021incorporating}. To better characterize the patterns of inter-trip time interval, we propose to use the Asymmetric Log-Laplace (ALL) distribution \citep{kotz2012laplace}, which is often used to model highly skewed data with peaks and heavy tails. The ALL distribution is specified by three parameters:
\begin{equation}\notag
    ALL\left(\tau; \beta, \lambda, \gamma \right) =
 \frac{\lambda \gamma}{\tau(\lambda + \gamma)} \begin{cases}
    \left(\frac{\tau}{\beta}\right)^\lambda &\text{if }  0<\tau < \beta, \\
    \left(\frac{\beta}{\tau}\right)^\gamma &\text{if }  \tau\geq \beta,
    \end{cases}
    \label{eq:alld}
\end{equation}
where the scale parameter $\beta$ controls the mode, and $\lambda$ and $\gamma$ are positive tail parameters for the left and right tails, respectively.

Given the high complexity and heterogeneity in individual travel behavior, a desirable distribution for modeling $p_{\theta}(\tau \mid \vec{h})$ should be able to approximate any general distributions arbitrarily well. Given that mixture models have the universal approximation (UA) property to approximate any probability density on $\mathbb{R}$ \citep{bishop1994mixture,dasgupta2008asymptotic,shchur2019intensity}, we use $\mathcal{D}$, as a mixture of ALL, to approximate $p_{\theta}(\tau \mid \vec{h})$:
\begin{equation}\label{eq:allm}
    \mathcal{D}\left(\tau; \vec{w},\vec{\beta}, \vec{\lambda}, \vec{\gamma} \right) = \sum^K_{k=1} w_k   ALL(\tau; \beta_k, \lambda_k, \gamma_k ),
\end{equation}
where $\vec{w}$ are the mixture weights. By using a mixture model, we can approximate the multi-modal travel patterns of an individual. For example, if a traveler only leaves records for morning commuting trips (e.g., take the metro in the morning but a taxi without records in the evening every day), we will observe a prominent peak at 24 hours. While for round-trip commuters, we expect $p\left(\tau | -\right)$ to have a peak around 8 hours (time at work) after a morning trip and having a peak around 16 hours (time at home) after an evening trip. Therefore, the mixture of ALL distributions provides a desired solution to learn $p\left(\tau | -\right)$ with multiple peaks. The logarithm of a variable under ALL mixture distribution follows the asymmetric Laplace mixture---$\operatorname{ALMixture}\left({{w}}, \hat{\vec{\beta}}, \hat{\vec{\lambda}}, \hat{\vec{\gamma}}\right)$, where each component is given by
\begin{equation}
\footnotesize
  AL\left(y\right) =\frac{\hat{\lambda}_k} {\hat{\gamma}_k + \frac{1}{\hat{\gamma}_k}} \begin{cases}
  \exp \left( \frac{\hat{\lambda}_k}{\hat{\gamma}_k} (y - \hat{\beta}_k)   \right) & \text{if }   0< y < \hat{\beta}_k, \\
     \exp \left( -\hat{\lambda}_k \hat{\gamma}_k (y - \hat{\beta}_k)   \right)& \text{if }   y \geq  \hat{\beta}_k,
    \end{cases}
\label{eq:trans}
\end{equation}
with $\hat{\beta}_k = \log(\beta_k)$, $\hat{\gamma}_k = \sqrt{\frac{\lambda_k}{\gamma_k}}$, and $\hat{\lambda}_k = \sqrt{{\lambda_k}{\gamma_k}}$.

Note that the log-likelihood in Eq.~\eqref{eq:trans} is easier to learn than that of the original ALL mixture model in Eq.~\eqref{eq:allm}. With this reparameterization, we use a special mixture density network (MDN) \citep{bishop1994mixture} to obtain the parameters in ALMixture:
\begin{equation}
\begin{split}
\vec{{w}}_n &= \operatorname{softmax}(\Phi_w \vec{h}_n + \vec{b}_w), \\ \hat{\vec{\beta}}_n &= \exp(\Phi_\beta \vec{h}_n + \vec{b}_\beta), \\
\vec{\hat{\lambda}}_n &= \exp(\Phi_\lambda \vec{h}_n + \vec{b}_\lambda), \\
\vec{\hat{\gamma}}_n &= \exp(\Phi_\gamma \vec{h}_n + \vec{b}_\gamma),
\end{split}
\end{equation}
where the $\operatorname{softmax}$ and $\exp$ transformations are applied to enforce the constraints on the distribution parameters, and $\{\Phi_w, \Phi_\beta, \Phi_\lambda, \Phi_\gamma, \vec{b}_w, \vec{b}_\beta, \vec{b}_\lambda, \vec{b}_\gamma\}$ are learnable parameters.

\subsection{OD Matrix Learning}

It is obvious that both origin $o$ and destination $d$ are correlated with time $t$ (or equivalently the trip lag $\tau$). Here we use a simple approach to model the conditional distribution $p^{*}(o_{n+1}\mid \tau_{n+1})$. Instead of directly modeling $p^{*}(o_{n+1}\mid \tau_{n+1})$, we make $o_{n+1}$ conditioned on the learned parameters $\{\vec{w}_n, \vec{\beta}_n, \vec{\lambda}_n, \vec{\gamma}_n \}$ of $\tau_{n+1}$, which helps us to avoid sampling data from the learned mixture distribution. Since the parameters of the ALL mixture have specific physical meaning, it is also simple to simulate trips from a learned model. We can modify some parameters of the learned ALL distribution to see the change of the origin and destination distribution. For example, we can reduce the values of the peak parameter $\vec{\beta}_n$ and evaluate the changes of origin-destination distribution. It can be regarded as an example of trip behavior change when an individual decides to travel earlier. Specifically, we have
\begin{align}
\begin{split}
    \vec{\hat{h}}_n &= \operatorname{concat}(\vec{{h}}_n, \vec{w}_n, \hat{\vec{\beta}}_n, \hat{\vec{\lambda}}_n, \hat{\vec{\gamma}}_n)  \\
    \hat{\vec{o}}_{n+1} &= \operatorname{softmax}(\Phi_o \hat{\vec{h}}_n + \vec{b}_o),
    \end{split}
\end{align}
the distribution $\hat{\vec{o}}_{n+1}$ for the next origin is dependent on the concatenated vector $\hat{\vec{h}}_n$, which is a combination of the history encoder $\vec{{h}}_n$ and the parameters of trip timestamps.

To model the destination distribution $p^{*}(d_{n+1})$, we multiply the origin output distribution $\hat{\vec{o}}_{n+1}$ with an OD matrix $O\!D_{n+1}$, in which each column stores the transition probability from one particular origin to all destinations (i.e., column sum is 1). Predicting individual's OD matrix has much broader applications than simply predicting the destination distribution $p^{*}(d_{n+1})$. For example, by assigning those OD matrices to every individual in a metro network, we can regulate each metro train's crowdedness. However, learning real-time OD matrices is extremely difficult due to the high-dimensionality ($S\times S$). To address this issue, we adopt the following approach to reduce the learning parameters in $O\!D_{n+1}$:
\begin{align}
    \begin{split}
        D^1_{n+1} &= \operatorname{reshape}(\Phi_{m}^1 \hat{\vec{h}}_n), \\
        D^2_{n+1} &= \operatorname{reshape}(\Phi_{m}^2 \hat{\vec{h}}_n), \\
        O\!D_{n+1} &= D^1_{n+1} {D^2_{n+1}}^{\top} \cdot M_{od}, \\
        O\!D_{n+1} &= \operatorname{softmax}(O\!D_{n+1}).\\
        \hat{\vec{d}}_{n+1} &= O\!D_{n+1}  \hat{\vec{o}}_{n+1}
    \end{split}
\end{align}
where $D^1_{n+1}, D^2_{n+1} \in \mathbb{R}^{S \times r}$, $S$ is the total number of stations in the transportation system, the size parameter $r$ is set to be much smaller than $S$ (i.e., $r \ll S$), $\Phi_{m}^1$ and $\Phi_{m}^2$ are learning parameters, $M_{od} \in \mathbb{R}^{S \times S}$ is applied to filter out impossible OD pairs by setting all unrelated pairs and OD pairs on the diagonal ($o=d$, not a meaningful trip) to $-\infty$, and the $\operatorname{softmax}$ operation is applied to keep the sum of each column in the OD matrix equals 1. Our model also captures dynamic OD dependencies, as it dynamically adjusts the weight of two connected nodes based on temporal encoding $\hat{\vec{h}}_n$. Moreover, the output destination distribution $\hat{\vec{d}}_{n+1}$, and origin distribution $\hat{\vec{o}}_{n+1}$ are connected with the learned parameters $[\vec{w}_n, \hat{\vec{\beta}}_n, \hat{\vec{\lambda}}_n$, and $\hat{\vec{\gamma}}_n]$. This strategy creates channels for information to flow between the initial output layers for event timing and the final layers for destinations, making AMTPP much easier to train.

\subsection{Model Training}

The parameters of AMTPP can be trained by the negative log-likelihood (NLL) of the inter-event times and origin-destination pairs in the training period. For learning, we use stochastic gradient descent. We pad the short trip sequence's time, origin and destination as the numbers of trips for users are different. The loss of the padding trips are masked. The final loss function for training AMTPP is given by
\begin{equation}
    \mathcal{L} =  - \sum_{u=1}^{U}\sum_{n=1}^{n_u}  \log p^*_{\Theta} (\tau^u_n) + \log p^*_{\Theta}(o^u_n)      + \log  p^*_{\Theta} (d^u_n),
\end{equation}
where $U$ is the total number of the users, $n_u$ is the total number of trips for individual $u$. For the log-likelihood of inter-trip time $\tau$, we apply a logarithm transformation on $\tau$ and obtain an asymmetric Laplace distribution for $y = \log(\tau)$, and finally we have the following log-likelihood for $\tau$:
\begin{equation}
\begin{split}
     \log p^*_{\Theta} (\tau) &= \log p^*_{\Theta}(y)  - \log (\tau),  \quad  y = \log(\tau),\\
     p^*_{\Theta}(y) &=  \operatorname{ALMixture}\left(\vec{{w}}, \vec{\hat{\beta}}, \vec{\hat{\lambda}}, \vec{\hat{\gamma}}\right).
    \end{split}
\end{equation}

\section{Experiments}
\subsection{Datasets and Baselines}
We evaluate AMTPP on two large-scale trip datasets:
\begin{itemize}
\item{\textbf{Guangzhou}:} This transit smart card dataset registers $\sim$650k trips from 10,000 anonymous users between July and September 2017 in Guangzhou metro system. There are $S=159$ metro stations in total. We use the data in the first two months (i.e., July and August) from 8,773 users for training and data from the rest users over the same period for validation. We evaluate/test model performance using trips in September. It should be noted that some users only appear in September (i.e., in the test period), and thus we have cold-start problem for these users.
\item{\textbf{Hangzhou}:} This dataset is also a  smart card dataset collected in Hangzhou metro with $S=81$ stations. It is released by TianChi\footnote{https://www.kaggle.com/zjplab/hangzhou-metro-traffic-prediction/activity} and publicly available. We conduct experiment on $\sim$600k trips from in January 2017 from 20,000 anonymous users, of which the trips over Jan 1 - Jan 20 are used for training and validation (18,000 users for training and the rest for validation). We assess model performance on trip data over the next five days (Jan 21 - Jan 25). 
\end{itemize}

We compare AMTPP with the following baselines---two simple models for location modeling and three neural TPPs:
\begin{itemize}
\item \textbf{Na\"ive}: We simply predict OD by reversing the OD of the previous trip, i.e., $o_{n+1}=d_n$ and $d_{n+1}=o_n$.
\item \textbf{Collective $T$-Gram:} A Bayesian n-gram model with regularized logistic regression \citep{zhao2018individual}. This work discretizes the time stamp into hours and assumes that users have trips every day. As this assumption does not hold for the two datasets, we only evaluate the prediction of origin and destination.
\item \textbf{RMTPP:} A TPP model that uses RNN to learn a representation of influences from past events. The intensity function follows a parametric form that is equivalent to the Gompertz distribution \citep{du2016recurrent}.
\item \textbf{FullyNN:} A feed-forward neural network is used to model cumulative intensity \citep{omi2019fully}.
\item \textbf{LNMix:} A TPP model that uses RNN to learn a representation of influences from past events, and a log-normal mixture model is used to model the conditional intensity function \citep{shchur2019intensity}.
\end{itemize}

\subsection{Experimental Setup}

We set the embedding sizes for origin, destination, hour of day and day of week to 64. For \textbf{Guangzhou}, we have access to additional
POI information (22 different types) which is pre-processed by inverse document frequency. We concatenate them into the origin and destination embeddings. For the self-attention layer, the number of heads for the attention model is set to 4, the size of the output hidden layer is set to 100. We use 16 asymmetrical log-Laplace components to model the distribution for $\tau$. The size parameter $r$ for the graph learning block is set to 3. We train AMTPP using Adam optimizer with an initial learning rate of 0.001. As the three neural TPP models are originally proposed for single marker, we adapt them by adding two multilayer perceptrons (MLPs) to directly output both $o$ and $d$.

\subsection{Results}

Table~\ref{tab:tod_comparison} compares the performance of AMTPP and baseline models for the prediction of time, origin, and destination on both \textbf{Guangzhou} and \textbf{Hangzhou} datasets. As can be seen, AMTPP achieves superior results on both datasets. All deep learning based TPP models outperform the Na\"ive and $T$-Gram baselines by a large margin. Compared to other neural TPP models, the proposed AMTPP achieves great improvement in predicting the distribution of timestamps $t$ and destinations $d$. In terms of predicting next trip time $t$, we find {FullyNN}, {LNMix} and AMTPP all outperform {RMTPP}, and AMTPP achieves substantial improvements over the two other baselines. Our results also suggest that the the asymmetrical log-Laplace mixture used in AMTPP is more representative than log-normal mixture in capturing the multimodal and peaked patterns in inter-trip time. Due to the lack of representation power, we find that the performance of FullyNN in predicting $t$ is as good as, if not better than, that of LNMix. In terms of destination prediction, AMTPP offers significantly better F1 score and accuracy, indicating that the proposed OD matrix learning block could capture the latent structural proximity among origin and destination. In terms of predicting the next origin $o$, we find it a rather simple task in which the Na\"ive baseline already achieves a high accuracy. While the improvements from all other models become marginal, AMTPP still obtains the best performance due to the introduction of the OD learning module.

\begin{figure}[!t]
\centering
\includegraphics[width=0.95\columnwidth]{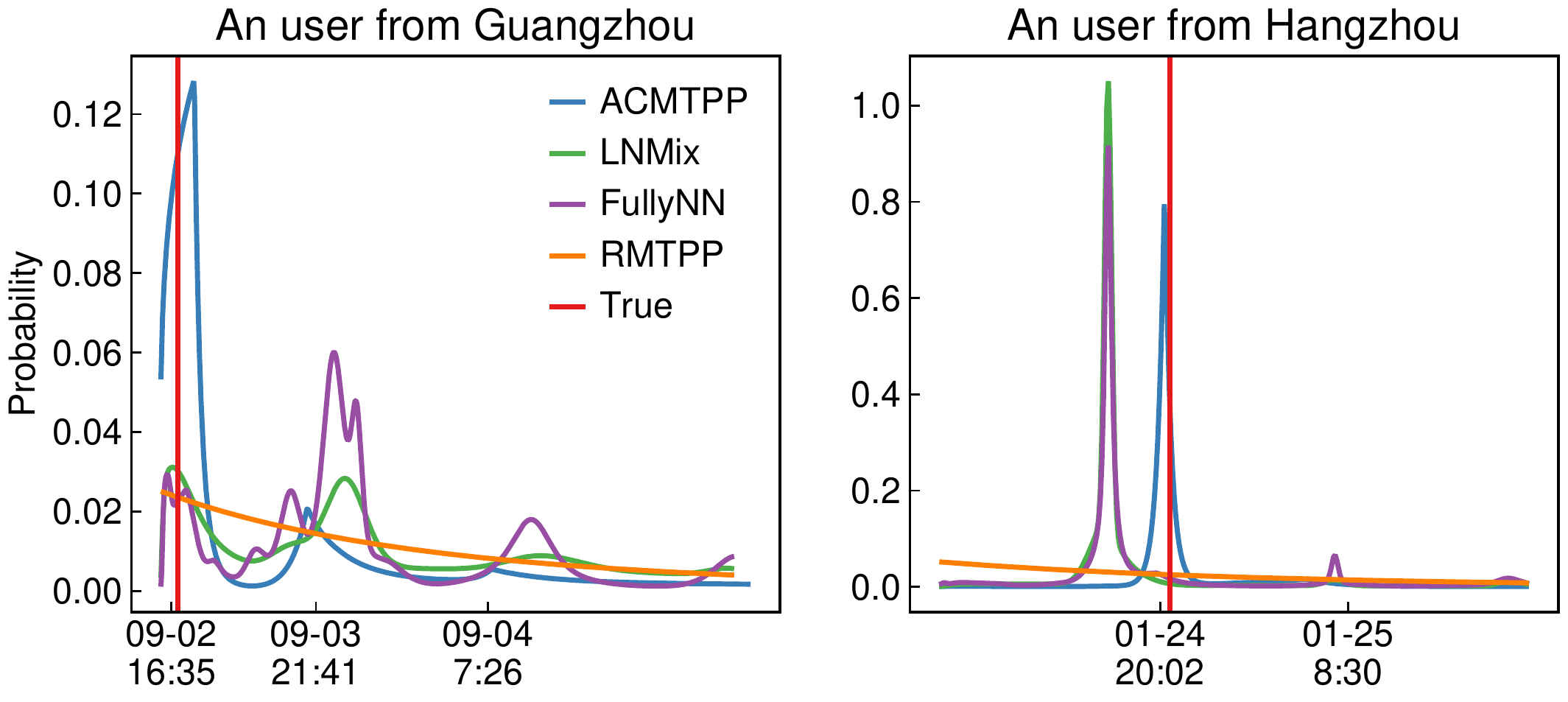}
\caption{Comparison of predicted trip departure time distribution of different TPP models. The vertical red lines show the ground truth. }
\label{Fig:3}
\end{figure}

\begin{table*}[!t]
  \centering
  \small
    \begin{tabular}{l|cc|cc|ccc|cc|cc|ccc}
    \toprule
    Dataset &\multicolumn{7}{c|}{Guangzhou (G)} & \multicolumn{7}{c}{Hangzhou (H)} \\
    \midrule
Model    & \multicolumn{2}{c|}{Accuracy} & \multicolumn{2}{c|}{F1-score} & \multicolumn{3}{c|}{NLL loss} & \multicolumn{2}{c|}{Accuracy} & \multicolumn{2}{c|}{F1-score} & \multicolumn{3}{c}{NLL loss} \\
          & {$o$} & {$d$} & {$o$} & {$d$} & {$t$} & {$o$} & {$d$} & {$o$} & {$d$} & {$o$} & {$d$} & {$t$} & {$o$} & {$d$} \\
          \midrule
    Na\"ive &0.713 &0.484 &0.706 &0.447 &/ &/  &/  &0.792 &0.643  &0.783  &0.641 &/  &/ &/  \\
    T-Gram &0.710 &0.481 &0.683 &0.449 &/ &4.606  &9.091  &0.794 &0.686  &0.785  &0.668 &/  &2.728 &5.731  \\
    RMTPP &0.725  &0.563  &0.722  &0.560  &4.103  &1.372  &2.143  &0.799  &0.708  &0.795  &0.706  &3.902  &1.001  &1.306  \\
    FullyNN &0.721  &0.561 &0.718  &0.557 &3.372 &1.388  &2.154  &0.797  &0.704  &0.793 &0.702  &2.864  &1.022  &1.337 \\
    LNMix & 0.721 &0.562 &0.719 &0.557  &3.507 &1.364  &2.127  &0.801  &0.709  &0.797  &0.708  &2.787  &0.982 &1.302  \\
    AMTPP &\textbf{0.750}  &\textbf{0.606} &\textbf{0.747} &\textbf{0.602} &\textbf{3.029}  &\textbf{1.261}  &\textbf{1.878}  &\textbf{0.805}  &\textbf{0.731}  &\textbf{0.805} &\textbf{0.732} &\textbf{2.535}  &\textbf{0.975} &\textbf{1.225}  \\
    \bottomrule
    \end{tabular}
\caption{Performance metrics for sequential prediction of trip attributes.}
  \label{tab:tod_comparison}%
\end{table*}%

Fig.~\ref{Fig:3} plots examples of estimated departure time distributions from different neural TPP models for two users in \textbf{Guangzhou} and \textbf{Hangzhou}, respectively. The user in Guangzhou data shows some random visits to different stations on a daily basis, while the user in Huangzhou has strong routine patterns---going to work and returning home every day. We can see that the intensity function of RMTPP is not suitable for modeling event data driven by human behavior---the simple Gompertz distribution cannot fully capture the temporal patterns of inter-trip time, which is strongly governed by the regularity and patterns in human behavior. FullyNN, LNMix and AMTPP can all reproduce the multimodal and peak patterns for both the random and regular users. However, due to the lack of knowledge of global time (e.g., time of day), FullyNN can LNMix sometimes cannot produce the correct peak (see the right panel for example). AMTPP, on the other hand, can benefit from it time-aware nature due to the introduction of time encoding, thus generating more accurate departure time distributions.

\begin{figure}[!t]
\centering
\includegraphics[width=0.85\columnwidth]{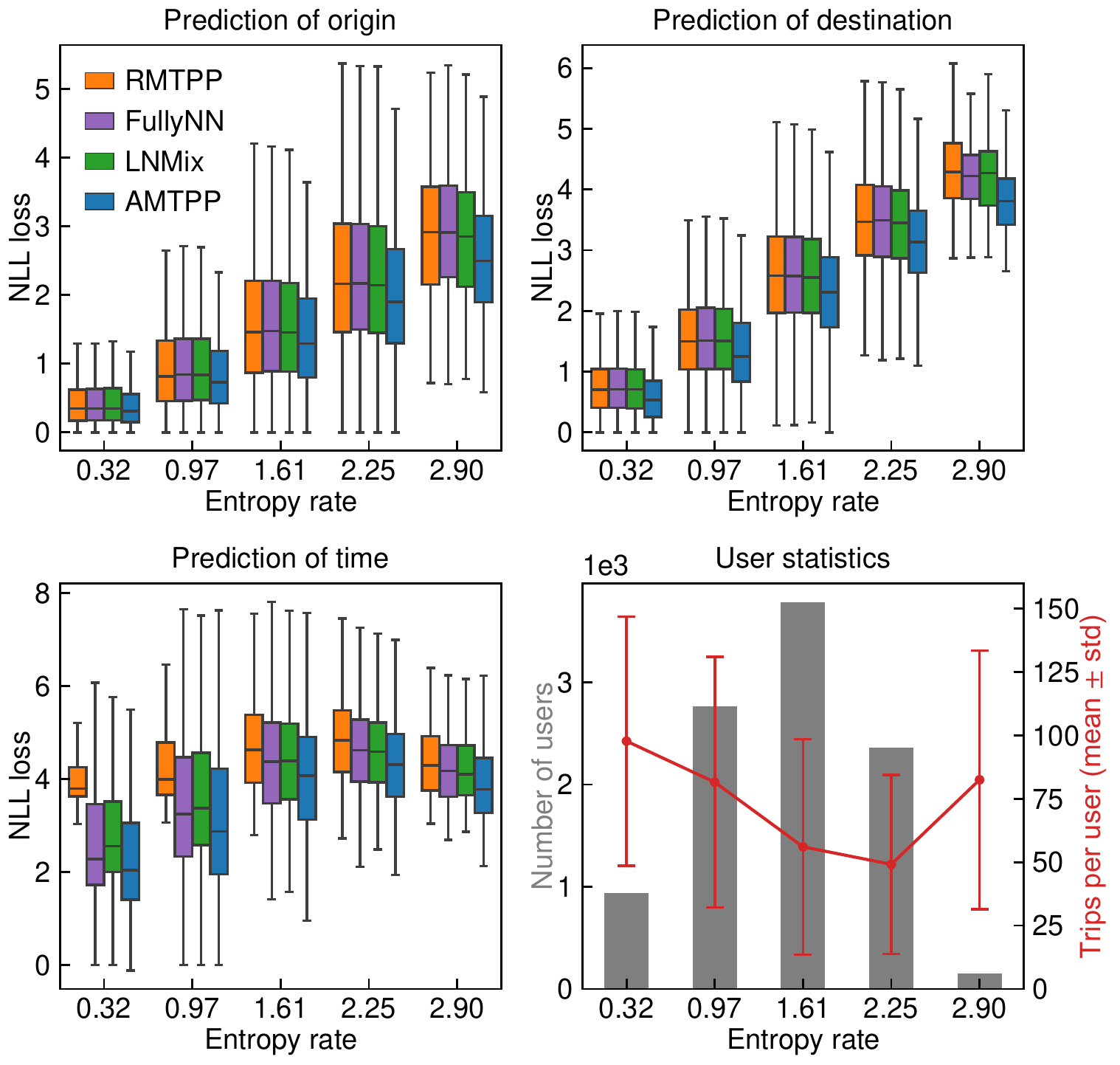}
\caption{Model performance for user groups with different levels of entropy rate (travel regularity).}
\label{fig:entropy_rate}
\end{figure}

We expect the prediction accuracy to be highly related to the degree of regularity in individual behavior. To empirically quantify the effect, we compare model performance for different user groups with diverse levels of behavior regularity. Based on \citet{goulet2017measuring}, we measure the travel regularity of user $u$ using entropy rate of his/her location sequence $[o_1^u, d_1^u, \cdots, o_{n_u}^u, d_{n_u}^u]$. The larger the entropy rate is, the more randomly a user travels, and the more difficult it becomes to predict the next trip. We divide users from the Guangzhou metro into five groups based on their entropy rate (see statistics in the bottom right panel of Fig.~\ref{fig:entropy_rate}). The first three panels in Fig.~\ref{fig:entropy_rate} show the NLL loss in predicting origin, destination, and time, respectively, for the five user groups. In general, we do observe the NLL loss of all the models in general increases with entropy rate. The proposed AMTPP has much lower NLL loss than other models in all prediction tasks and across different levels of entropy rate, suggesting that AMTPP universally improves the mobility prediction of different types of users. 

\subsection{Ablation Study}

We next conduct ablation study to assess the effectiveness of self-attention mechanism, ALL mixture distribution, and OD matrix learning block. We quantify NLLs of different versions/configurations of AMTPP on  \textbf{Guangzhou}: 1) LogNormMix $t$, which replaces the asymmetrical log-Laplace mixture with log-normal mixture; 2) No OD matrix, which uses two MLPs to directly output $o$ and $d$; 3) No time embedding, which removes the embeddings in Eq.~\eqref{eq:time_embedding}; 4) Fixed embedding, which uses a fixed frequency $L_{pos} = 10000$ instead of making it learnable. Table~\ref{tab:ablation} presents the results of the ablation study. As we can see, all the designed features improve model performance and enhance the approximation capacity of AMTPP. In particular, we find the introduction of time embeddings and the use of ALL mixture play a critical role in modeling individual mobility. This further confirms the importance of time awareness and activity duration in modeling human behavior with strong rhythmic and periodic patterns. Comparing the results in Table~\ref{tab:tod_comparison}, we can see AMTPP with no time embeddings still outperforms other neural TPPs, indicating the superiority of using ALL mixture. In addition, we can see that the OD learning block and learnable frequency also improves the accuracy of origin and destination prediction.

\begin{table}[!t]
\small
\centering
\begin{tabular}{lrrr}
\toprule
Configuration & NLL $t$ & NLL $o$ & NLL $d$ \\
\midrule
 LogNormMix $t$ &3.150  &1.302 &1.912  \\
 No OD matrix &3.045  &1.321 &1.988 \\
 No time embedding &3.236 &1.420 &2.112 \\
 Fixed embedding &3.051 &1.360 &2.051 \\
 AMTPP &3.029  &1.261  &1.878  \\
\bottomrule
\end{tabular}
\caption{Ablation results for different configurations.}
\label{tab:ablation}
\end{table}

\section{Conclusion}

In this paper, we present a novel TPP framework for modeling multidimensional timestamped trip records $(t,o,d)$ and predicting individual mobility. In particular, we develop AMTPP that considers both origin and destination as coupled markers and uses two strategies to effectively model the rhythms and regularity in travel behavior. The first is to introduce time embedding to make the model time-aware and the second is to characterize the multimodal and peaked inter-trip time distribution using and ALL mixture. In addition, we also propose an effective method to learn the hidden relationship between origin and destination dynamically. We compare AMTPP with other neural TPP models on smart card-based metro trip datasets from two large cities, and our experiment results show that the proposed AMTPP achieves state-of-the-art performance. Overall, AMTPP opens new directions to model human behavior using TPPs. For future work, we plan to examine other time embedding techniques such as \citet{kazemi2019time2vec}, extend it to model systems with higher-order markers (e.g., including activity type), and use the learned model to generate synthetic data for agent-based simulations \citep{w2016multi}.

\clearpage

\bibliographystyle{aaai22}
\bibliography{ref}

\end{document}